\documentclass{llncs}
\usepackage[utf8]{inputenc}
\usepackage[backend=biber]{biblatex}
\usepackage{graphicx}
\usepackage{color}
\usepackage[hidelinks]{hyperref}
\usepackage{booktabs}
\usepackage{orcidlink}
\usepackage{todonotes}
\usepackage[frozencache]{minted}
\usepackage{paralist}
\usepackage{newdrawline}
\usepackage{fca}
\usepackage{cleveref}
\usepackage{xurl}
\hypersetup{breaklinks=true}
\let\cref\Cref

\bibliography{bibliography.bib}

\definecolor{rja}{rgb}{0.878, 0.831, 0.482} 
\definecolor{tha}{rgb}{0.721, 0.576, 0.862} 

\definecolor{TODO}{rgb}{0.784,0.145,0.00}

\begin{document}

\title{A Repository for Formal Contexts}

\author{
  Tom Hanika\inst{1}\orcidlink{0000-0002-4918-6374}
  \and
  Robert Jäschke\inst{2}\orcidlink{0000-0003-3271-9653}
}

\institute{
  University of Hildesheim\\
  \email{tom.hanika@uni-hildesheim.de}
  \and
  Berlin School for Library and Information Science\\ Humboldt-Universität zu Berlin \\
  \email{robert.jaeschke@hu-berlin.de}
}

\maketitle

\begin{abstract}
  Data is always at the center of the theoretical development and
  investigation of the applicability of formal concept analysis. It is
  therefore not surprising that a large number of data sets are
  repeatedly used in scholarly articles and software tools, acting as
  de facto standard data sets. However, the distribution of the data
  sets poses a problem for the sustainable development of the research
  field. There is a lack of a central location that provides and
  describes FCA data sets and links them to already known analysis
  results. This article analyses the current state of the
  dissemination of FCA data sets, presents the requirements for a
  central FCA repository, and highlights the challenges for this.
\end{abstract}



\section{Introduction}\label{sec:introduction}
Reproducing and comparing results are mandatory for the scientific method. This is in particular true for data-driven and data-centered research fields, such as formal concept analysis (FCA). Research data repositories are therefore an integral part of a functioning research ecosystem.

A large number of such repositories have been established in the field
of machine learning, such as the \emph{The UCI Machine Learning
  Repository}~\cite{uci}, the platform
OpenML~\cite{vanschoren2014openml}, or  \emph{Hugging
  Face}.\footnote{\url{https://huggingface.co/}} There is a
distinction between general and specialized repositories.  The latter
can be specialized, for example, in data types, application areas, or
learning methods. Without such repositories, freely circulating data
sets are used in a research community. It is often not possible to
trace where the data originates from, how it may have been derived
from another data set, whether it is complete and correct, etc. What's
more, data can disappear from the community and no longer be found,
for example, if researchers retire or websites vanish.

This problem and its effects increasingly affect the research field of
FCA, which has also been recognized
previously~\cite{andrews2009conversion,borchmann2016experimental,orphanides2013fcawarehouse}.
Various researchers have approached this problem in part by providing
websites\footnote{for example, \url{https://upriss.github.io/fca/examples.html}}
for the FCA community that link different sources of information, or
by including test data
sets\footnote{\url{https://github.com/tomhanika/conexp-clj/tree/dev/testing-data},
  \url{https://github.com/neuroimaginador/fcaR/tree/master/data}} in
their FCA software~\cite{DBLP:conf/icfca/HanikaH19,cordero2022fcar}.

A problem analysis and initial approaches to solving this issue have
already been described independently by
S. Andrews~\cite{andrews2009conversion} and C. Orphanides and
G. Georgiou~\cite{orphanides2013fcawarehouse}. More than ten years
later, however, none of the proposed solutions have been manifested
themselves in real existing repositories. This is where this work
comes in by
\begin{inparaenum}[a)]
\item presenting an updated analysis of the requirements for an FCA
  repository,
\item describing an approach for an initial solution, and
\item presenting a rudimentary
  repository,\footnote{\url{https://fcarepository.org/}} that we have
  already initiated, and its future development.
\end{inparaenum}

With our solution we follow the KISS\footnote{\emph{keep it simple,
    stupid}} principle, favoring a simple functioning repository (with
few features) over a complicated and therefore difficult to implement
repository. Already in its present form, the repository allows for
\begin{inparaenum}[a)]
\item testing algorithms (and implementations) on correctness,
\item benchmarking algorithms,
\item comparing different FCA procedures,
\item helping beginners to explore or learn FCA.
\end{inparaenum}
It also contains the ``classic'' data sets from Ganter and Wille's FCA
book~\cite{ganter1999formal}.  Furthermore, our approach supports
low-threshold access similarly to other popular data science
libraries, for example,
\emph{Seaborn}\footnote{\url{https://github.com/mwaskom/seaborn-data}}
via its \verb|load_dataset()| method.
In our repository located at \url{https://fcarepository.org/},
\begin{inparaenum}[a)]
\item each context is a file in a git repository, and
\item the metadata for each context is described in a file that is
  machine-readable and human-editable.
\end{inparaenum}

With this modeling we obtain version control, a workflow for
collaboration and contributions (forks, pull requests), a continuous
integration pipeline for the automatic generation of derivatives
(e.g., human-readable documentation, statistics, lattice diagrams),
simple programmatic access using HTTP, etc. Therefore, the repository
could easily be integrated into FCA workflows, tools, and libraries
and can enable and simplify the (re)use of FCA data. For
sustainability, we envision to create snapshots of the git repository
on a regular basis and to publish them on suitable platforms, such as
Zenodo.\footnote{\url{https://zenodo.org/}}

In order to make the FCA repository sustainable and as scientifically
reliable as possible, many challenges still need to be
overcome. Specifically:
\begin{inparaenum}[a)]
\item a curation policy and metadata schema has to be developed,
\item more formal contexts and metadata have to be collected, and
\item authors of FCA tools and libraries need to implement a reliable interface to the repository.
\end{inparaenum}
We hope that our initial analysis and solution helps to raise awareness for the repository problem and starts the discussion that gathers feedback from the FCA community.

This paper is organized as follows: In Section~\ref{sec:related} we
give an overview on related work and then discuss existing related
approaches in more depth in Section~\ref{sec:existing}. In the main
part in Section~\ref{sec:method} we analyze requirements for an FCA
repository and provide suggestions for implementation. Finally, we
discuss next steps and the organization of the repository in
Section~\ref{sec:organisation} and conclude the paper with a
discussion in Section~\ref{sec:discussion}.


\section{Related Work}\label{sec:related}

There exists a wide variety of repositories that host research data --
institutional, domain-specific, and generic, such as
Zenodo~\cite{zenodo}, which was launched by the
OpenAIRE\footnote{\url{https://www.openaire.eu/}} partner CERN. Most
of them are clearly intended as archives for research data without
providing any insights into the data apart from basic metadata (e.g.,
license, authorship, file format). Registries, such as, \emph{re3data}\footnote{\url{https://re3data.org/}} record more than 3200 different repositories~\cite{re3}.

A prominent example in computer science is the UCI Machine Learning
Repository~\cite{uci}. Founded in 1987, it is one of the oldest
repositories hosting data sets for the empirical analysis of machine
learning algorithms. With a focus on tabular data of `instances'
described by `variables' it provides basic information about each
variable (e.g., role, type, unit, description).
In the realm of machine learning, OpenML~\cite{vanschoren2014openml}
considerably extends this idea by allowing everyone to share data sets,
tasks, implementations, and results. Its goal is to enable ``networked
science'' that uses ``online tools to share, structure and analyse
scientific data on a global scale'' \cite{vanschoren2014openml}.

KONECT, the Koblenz Network Collection \cite{kunegis2013konect}, is a
repository dedicated to network science (i.e., graph-represented) data
sets. Its web site facilitates the exploration of such data sets by
providing more than thirty descriptive network statistics (e.g.,
average degree, mean distance) and data visualizations (e.g., spectral
distributions, degree assortativity). In addition, a handbook and a
toolbox for the statistics framework GNU/R supports researchers in
using the data.

Finally, DraCor~\cite{fischer2019programmable} is an example from the
humanities, specifically literature. It is a website and REST-based
API that provides multi-faceted access to drama corpora, coined
``programmable corpora'' by its founders. Apart from the full-texts of
the dramas, DraCor provides metadata, and structured information
(e.g., speaker text) together with a web-based interface for
exploration of the character co-occurrence networks.

Various trends can be identified in the recent development of research
data repositories. On the one hand, the creation and further
development of existing generalist repositories. Secondly, the
creation of domain-specific repositories, for example, for
agricultural science~\cite{specka2023fairagro} or
mathematics~\cite{danabalan2023mardi}, or application-specific, for
instance, collaborative research
centers~\cite{10.1093/gigascience/giad049}. The latter two are
particularly favored in Germany by an initiative for a national
research data infrastructure.

There are reasons for the emergence and further development of
repositories that are specifically tailored to certain areas of
research: by focusing on certain types of data and file formats, they
can provide dedicated services, for example, additional metadata,
faceted search, an integration into the tools and workflows of the
research community, or draw on the expertise from other domain
experts.


A web-based repository for formal contexts was first described in 2009
by S.~Andrews~\cite{andrews2009conversion}. The idea was that the
repository hosts donated and randomly-generated data sets, ``along
with, where possible, [their] original data file, information about
the original data (probably from the original data source), a link
back to the original data source, the conversion log, and FCA
information, such as context density and number of concepts''
\cite{andrews2009conversion}. However, few details were provided and
-- to the best of our knowledge -- the repository either was never
realized or is no longer available online. Four years later
C.~Orphanides and G.~Georgio proposed
\emph{FCAWarehouse}~\cite{orphanides2013fcawarehouse}, a website no
longer in service.

In 2016, Hanika and Borchmann \cite{borchmann2016experimental}
reiterated the need for an archive of data and computational results
related to FCA when they uncovered implausible results in the
Stegosaurus phenomenon. As far as the authors are aware, there is no
other domain-specific FCA repository apart from the one presented
here.



\section{Existing Related Approaches}\label{sec:existing}
As described in the previous section, there are no other FCA
repositories. However, there are a number of tools and other sources
of information that perform some of the tasks of an FCA
repository. Moreover, the authors are keen to integrate the proposed
repository into existing FCA tools. We therefore provide a brief
overview for them. In addition, we can justify some of the design
decisions of the FCA repository based on their characteristics.

\subsection{Tools}\label{sec:fca-tools}
There is an abundance of software libraries and tools to conduct
formal concept analysis. In the following we provide a brief overview
on the most recent tools. Therefore, we omitted offline-tools that
were not updated for more than two years. Moreover we discarded for
our analysis pure concept mining tools, such as \emph{pcbo} or
\emph{inclose}. We acknowledge that the overview is not exhaustive and
might miss important tools.

\begin{inparadesc}
\item[xflr6 / concepts] is a basic but useful Python3 library for
  FCA. It allows for computing and drawing concept
  lattices.\footnote{\url{https://github.com/xflr6/concepts}} Binary comma separated value (CSV)
   as well as the Burmeister format are supported for loading and
  storing formal contexts.
\item[FCApy] is also a Python3 FCA library with a focus on interacting
  with machine learning algorithms.
\item[fcaR] is a popular library within the GNU/R ecosystem. In
  particular it allows for computing implications and conceptual
  scaling~\cite{cordero2022fcar}
\item[ConExp] -- ``The Concept Explorer'' is one of the classic FCA
  tools. Although its last release was in 2013, it is still very
  popular due to its intiuitive graphical user interface.
\item[FcaKit] is a software library written in the Swift programming
  language, which is predominantly used in the iOS / MacOS
  ecosystem. Many concept mining and factorization algorithms are
  implemented within this library.
\item[conexp-clj] is one of the most versatile software libraries for
  FCA. It is implemented in the functional programming language
  Clojure and allows interaction with Java code, among other things.
\item[FCA Tools Bundle] is an online
  platform~\cite{DBLP:conf/icfca/CristeaSS19} and has a large
  collection (i.e., 166 files) of formal contexts. The context are
  presented with minor meta data (e.g., number of objects, etc) but
  only minimal provenience information is available, as many data sets
  have no description beyond one or a couple of words. Some data sets
  can be exported as CSV files and others using the Burmeister
  format. The focus of this platform is on the analysis of triadic and
  polyadic contexts.
\item[LatViz] is also an online platform which allows editing formal
  context and computing their concept lattice. Data sets can only be
  exchanged via an undocumented JSON format. However, a converter from
  binary CSV is provided as an offline tool.
\end{inparadesc}
All these tools and their attributes are depicted
in~\cref{fig:ctx}. The corresponding lattice is shown
in~\cref{fig:lat}. For a more exhaustive and extensive comparison of
FCA tools we refer the reader to Saab et
al.~\cite{saab2022evaluating}. Their analysis includes many tools that
we have discarded for the reasons mentioned above.

\begin{figure}[t]
  \centering
    \input{tools.tex}
    \caption{Comparison of recent FCA tools. The classic Conexp tool
      was included for comparison. Also non-recent online plattforms
      were added.}
  \label{fig:ctx}
\end{figure}

T.~Tilley already addressed the problem of tool support in
2004~\cite{tilley2004tool} and U.~Priss addressed the problem of data
interoperability for FCA in
2008~\cite{priss2008fca,priss2008fcastone}. Unfortunately, none of the
tools U.~Priss considered in her work made it into our comparison, as
there have been no releases for many years. Irrespective of this, the
considerations made are still valid. In particular, the finding that
the Burmeister format and the representation as a binary CSV have the
widest support among the FCA tools.

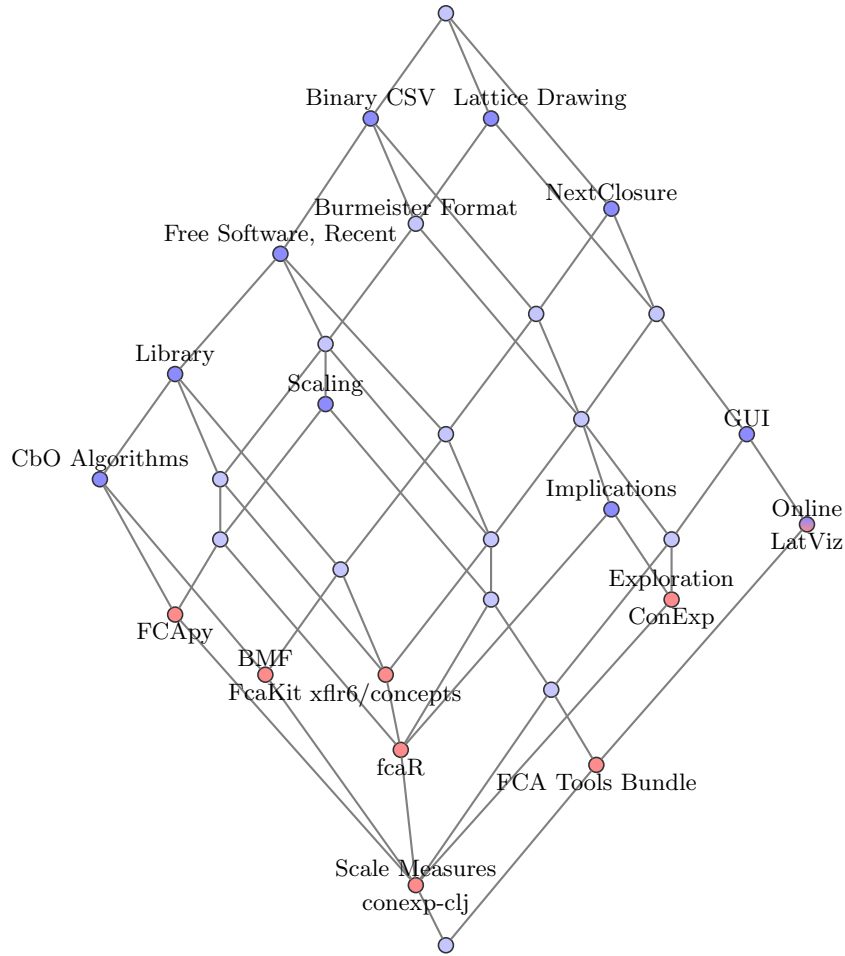
\begin{figure}[t]
  \centering
  \colorlet{mivertexcolor}{blue}
\colorlet{jivertexcolor}{red}
\colorlet{vertexcolor}{mivertexcolor!50}
\colorlet{bordercolor}{black!80}
\colorlet{linecolor}{gray}
\tikzset{vertexbase/.style 2 args={semithick, shape=circle, inner sep=2pt, outer sep=0pt, draw=bordercolor},%
  vertex/.style 2 args={vertexbase={#1}{}, fill=vertexcolor!45},%
  mivertex/.style 2 args={vertexbase={#1}{}, fill=mivertexcolor!45},%
  jivertex/.style 2 args={vertexbase={#1}{}, fill=jivertexcolor!45},%
  divertex/.style 2 args={vertexbase={#1}{}, top color=mivertexcolor!45, bottom color=jivertexcolor!45},%
  conn/.style={-, thick, color=linecolor}%
}
\begin{tikzpicture}
  \begin{scope}[scale=0.2] 
    \begin{scope} 
      \foreach \nodename/\nodetype/\param/\xpos/\ypos in {%
        0/vertex//0.0/0.0,
        1/jivertex//-2.0/4.0,
        2/jivertex//10.0/12.0,
        3/jivertex//-3.0/13.0,
        4/vertex//7.0/17.0,
        5/jivertex//-12.0/18.0,
        6/jivertex//-4.0/18.0,
        7/jivertex//-18.0/22.0,
        8/vertex//3.0/23.0,
        9/jivertex//15.0/23.0,
        10/vertex//-7.0/25.0,
        11/vertex//-15.0/27.0,
        12/vertex//3.0/27.0,
        13/vertex//15.0/27.0,
        14/divertex//24.0/28.0,
        15/mivertex//11.0/29.0,
        16/mivertex//-23.0/31.0,
        17/vertex//-15.0/31.0,
        18/vertex//0.0/34.0,
        19/mivertex//20.0/34.0,
        20/vertex//9.0/35.0,
        21/mivertex//-8.0/36.0,
        22/mivertex//-18.0/38.0,
        23/vertex//-8.0/40.0,
        24/vertex//6.0/42.0,
        25/vertex//14.0/42.0,
        26/mivertex//-11.0/46.0,
        27/vertex//-2.0/48.0,
        28/mivertex//11.0/49.0,
        29/mivertex//-5.0/55.0,
        30/mivertex//3.0/55.0,
        31/vertex//0.0/62.0
      } \node[\nodetype={\param}{}] (\nodename) at (\xpos, \ypos) {};
    \end{scope}
    \begin{scope} 
      \path (11) edge[conn] (17);
      \path (11) edge[conn] (21);
      \path (6) edge[conn] (17);
      \path (6) edge[conn] (10);
      \path (6) edge[conn] (12);
      \path (22) edge[conn] (26);
      \path (3) edge[conn] (11);
      \path (3) edge[conn] (6);
      \path (3) edge[conn] (15);
      \path (3) edge[conn] (8);
      \path (19) edge[conn] (25);
      \path (9) edge[conn] (15);
      \path (9) edge[conn] (13);
      \path (29) edge[conn] (31);
      \path (4) edge[conn] (8);
      \path (4) edge[conn] (13);
      \path (17) edge[conn] (22);
      \path (17) edge[conn] (23);
      \path (0) edge[conn] (1);
      \path (0) edge[conn] (2);
      \path (27) edge[conn] (29);
      \path (27) edge[conn] (30);
      \path (28) edge[conn] (31);
      \path (10) edge[conn] (22);
      \path (10) edge[conn] (18);
      \path (16) edge[conn] (22);
      \path (30) edge[conn] (31);
      \path (1) edge[conn] (3);
      \path (1) edge[conn] (9);
      \path (1) edge[conn] (4);
      \path (1) edge[conn] (5);
      \path (1) edge[conn] (7);
      \path (5) edge[conn] (10);
      \path (5) edge[conn] (16);
      \path (15) edge[conn] (20);
      \path (26) edge[conn] (29);
      \path (20) edge[conn] (27);
      \path (20) edge[conn] (25);
      \path (20) edge[conn] (24);
      \path (14) edge[conn] (19);
      \path (23) edge[conn] (27);
      \path (23) edge[conn] (26);
      \path (21) edge[conn] (23);
      \path (8) edge[conn] (21);
      \path (8) edge[conn] (12);
      \path (25) edge[conn] (28);
      \path (25) edge[conn] (30);
      \path (12) edge[conn] (20);
      \path (12) edge[conn] (23);
      \path (12) edge[conn] (18);
      \path (7) edge[conn] (11);
      \path (7) edge[conn] (16);
      \path (2) edge[conn] (4);
      \path (2) edge[conn] (14);
      \path (13) edge[conn] (19);
      \path (13) edge[conn] (20);
      \path (18) edge[conn] (26);
      \path (18) edge[conn] (24);
      \path (24) edge[conn] (29);
      \path (24) edge[conn] (28);
    \end{scope}
    \begin{scope} 
      \foreach \nodename/\labelpos/\labelopts/\labelcontent in {%
        1/above//{Scale Measures},
        5/above//{BMF},
        9/above//{Exploration},
        14/above//{Online},
        15/above//{Implications},
        16/above//{CbO Algorithms},
        19/above//{GUI},
        21/above//{Scaling},
        22/above//{Library},
        26/above//{Free Software, Recent},
        27/above//{Burmeister Format},
        28/above//{NextClosure},
        29/above//{Binary CSV},
        30/above//{\hspace{4em}Lattice Drawing},
        1/below//{conexp-clj},
        2/below//{FCA Tools Bundle},
        3/below//{fcaR},
        5/below//{FcaKit},
        6/below//{ xflr6/concepts},
        7/below//{FCApy},
        9/below//{ConExp},
        14/below//{LatViz}
      } \coordinate[label={[\labelopts]\labelpos:{\labelcontent}}](c) at (\nodename);
    \end{scope}
  \end{scope}
\end{tikzpicture}
  \caption{Concept Lattice for the formal context in~\cref{fig:ctx}.}
  \label{fig:lat}
\end{figure}

\subsection{Data Collections}

U.~Priss has some `classic' contexts on her web
page\footnote{\url{https://upriss.github.io/fca/examples.html}} and
some FCA tools have contexts for unit tests (e.g.,
conexp-clj\footnote{\url{https://github.com/tomhanika/conexp-clj/tree/dev/testing-data}}
or
concepts\footnote{\url{https://github.com/xflr6/concepts/tree/master/examples}})
but these are neither comprehensive nor easy to find, they have no
machine-readable metadata, they are not integrated into FCA tools or
libraries, and they are sometimes difficult to cite.

\section{Analysis and Proposition}\label{sec:method}
In this section, we analyze the requirements and the corresponding
simple solutions for the planned FCA repository in detail. The goal of
supporting the FCA community with the repository should guide us in
all decisions. It is clear to us that the attempt to create a
comprehensive FCA platform is doomed to failure and will result in
software that resembles a jack-of-all-trades. Thus, the overall
guiding principle will be KISS.

\subsection{Parts}

The main parts of the planned FCA repository are briefly listed below
and their necessity is explained in short.

\paragraph{Contexts:} The central entity is the formal context. All
parts should be depicted explicitly, that is, objects, attributes, and incidence relation.

\paragraph{Simple Statistics, Metadata and Usage:} For every context
simple statistics, for example, number of objects, density, etc. shall
be provided.
Moreover, metadata including provenience, contributor, editor etc. is
necessary. Also, the fact if a formal context is artificial or derived
from real data should be noted. In addition, an overview of where a
context has already been used/analyzed, for instance, in which
scholarly articles, would be helpful.

\paragraph{Relations:} A central element of FCA is control over
scaling. Since many formal contexts are derived from non-binary data,
access to the scales used is essential. Furthermore, sub-contexts are
often used for analyses. This information is also important for the
analyst. These and other relationships between formal contexts are to
be mapped by the FCA Repo.

\paragraph{Collections:} We envision that standard collections of
formal contexts can be useful. For example, when evaluating a new
algorithm one may employ a \emph{standard benchmark set of
  contexts}. These and similar tasks require the compilation and
labeling of named collections of formal contexts.

\paragraph{Concepts, Lattices, Diagrams:} Although not essential, it
would be very helpful to store the formal concepts belonging to a
formal context, the concept lattice, or even a lattice diagram. This
would increase the scientific reproducibility of results and the
ecological sustainability of analyses.

\paragraph{Implication Bases:} Similarly to the last point, it would
be very helpful to store implication bases.

\subsection{Implementation Considerations}\label{sec:implementation}

Implementing the envisioned FCA repository requires storing and
curating formal contexts, possibly additional data, and descriptive
metadata. In this section, we discuss aspects that need to be taken
into account and make specific suggestions for implementing the
repository.
As an overall pre-condition, we restrict ourselves to a
\emph{file-based} repository, as files are the typical format in which
formal contexts are stored and used.

\subsubsection{Files} A file in a repository is essentially
characterized by its \emph{name}, its \emph{location}, and its
\emph{content}.

%
%
\paragraph{Name.} The name of a file acts as an identifier within a
file system and must be unique at least within one directory. In
addition, the
file name acts as a visible identifier to everyone who is using the
file. We consider two approaches for naming a file: choosing a
meaningful name or simply using an arbitrary identifier (e.g., a
number or UUID \cite{rfc4122}) as name. On the one hand, a meaningful
name has some benefits for humans, for example, the name can give an
indication to a file's content. On the other hand, choosing a good
name is a hard problem~\cite{Benner2023}, as the following
Example~\ref{ex:filename} shows.
%
\begin{example}\label{ex:filename}
  Let us consider the formal context from Figure~1.1 of the FCA
  book~\cite{ganter1999formal}. The caption states that the context is
  from an educational film ``Living Beings and Water'', so a
  meaningful file name could be based on the title of that film. When
  creating a file name from that title, immediately certain questions
  arise: How many words to include? How should words be separated (by
  space, underscore, or not at all)? Where and how to use lower/upper
  case characters (all lower/upper case, title capitalization, or
  CamelCase)? And, given that the book is the English translation of
  the 1996 German original \cite{ganter1996formal} and the film's
  original (German) title is ``Lebewesen und Wasser'', we should
  consider which language to use: the original language or English as
  a common and default language.
\end{example}
Furthermore, the file name could also include metadata to provide more
information to users, for example, an ISO language code \cite{iso639}
to signal the language of the object and attribute names of the
context and to distinguish it from translated variants of the same
context. Typically, the file name also includes a file extension which
indicates the file format and thus the structure of the content.
Despite the challenges, we propose to use a meaningful file name based
on the content of the formal context. Our rationale is, that the
repository is intended for formal contexts (i.e., files) to be downloaded
and used by researchers. And in that scenario the file name should
bear some meaning to the user, as they have to store and find the file
on their computer's file system.
The exact format of the file name is up to discussion, but for now we
suggest to keep it short, all lowercase, in English, words separated
by underscore, and appending the ISO language code (to allow for
distinguishing translations of contexts).
We are aware that naming things comes with some power, since together
with their repository URL the file names can easily become unique
identifiers for the corresponding contexts. Later on, we propose a
curation policy that shall ensure conscientious naming.

\paragraph{Location.} The location of a file within the repository
needs to be specified. This is partly a matter of the overall
structure of the repository, that is, how it is organized into
directories and subdirectories.
We propose to have one directory, named \texttt{contexts}, in the
top-level directory of the repository that contains the files for all
contexts. Additional files, that, for example, contain lattices or
implication bases can be put into suitably named additional
directories.
Together with the base URL of the repository, the \texttt{contexts}
directory and the file name specify a URL which acts as a unique
identifier (and locator) for each context.



\paragraph{Content.} The content of the file can be represented in
various file formats \cite{priss2008fca,priss2008fcastone}. A
discussion of their benefits and drawbacks is beyond the scope of this
work, see U.~Priss~\cite{priss2008fcastone} for an overview. Since
FCAStone \cite{priss2008fcastone} can convert between some of the more
common formats, we propose to use the format introduced in ConImp by
Peter Burmeister \cite{burmeister2003conimp} as default. As a plain
text format, it is easy to understand for humans but also easy to
parse for machines. As we have seen in Section~\ref{sec:fca-tools}, it
is also well supported by still maintained tools and libraries.

Files in that format are typically idenfied by the extension ``cxt''
(cf. \cite[footnote~20]{burmeister2003conimp}).
Since, within the specification of the format, the names and order of
objects and attributes can be freely chosen, we propose to stay as
close to the original source as possible.
Furthermore, we propose to use UTF-8 as text encoding for names of
objects and attributes. Although at the time of conception some older tools may have only supported ASCII characters, we believe the repository should be geared towards current and future tools and use cases.
Additional files formats can be provided for contexts. Since a
conversion could easily be automated, we can also imagine a conversion
service for the repository. Nevertheless, the Burmeister (``cxt'')
format should be the gold standard.

\subsubsection{Metadata}

%
%

Metadata shall be provided for all contexts, as it provides context
and important information for humans, can simplify processing of the
data, and is crucial for applications built on top of the repository
(for example, an exploratory web page). These aspects are particularly
important with regard to the scholarly use of the FCA repository.
We discuss the \emph{what}, \emph{how}, and \emph{where}.

\paragraph{What.} Contexts typically have a \emph{title} which is often
used to refer to them (e.g., ``Living Beings'').
At least the title but typically also the names of objects and
attributes are in a certain \emph{language}.
Since the repository should host well-known, published contexts, the
\emph{source} where the context was published or used should be
provided. What is considered to be the source is an open question --
should it be the first (published) use in formal concept analysis or
the first publication of the data at all?\footnote{Or something more
  complex, for example, comprising the citation chain from the first
  publication of the data to the first published use in FCA.}
A \emph{description} should provide further information, for example,
the meaning of the attributes or how the context was constructed.
Except \emph{language}, we consider all of these properties to be
mandatory.
If the \emph{language} is not English, it must be specified.
Other metadata is conceivable, for example, to model relationships
between contexts (e.g., derived  or translated contexts).

\paragraph{How.} To store the metadata, a suitable data representation
together with a serialization into a file is required.
On the one hand, the amount and complexity of the metadata is
tractable; on the other hand, the representation should not impose too
many restrictions but support later extensions. Thus, a good trade-off
between expressivity and simplicity is required.

Having a look at the proposed metadata fields for each context,
\emph{title}, \emph{description}, \emph{language} could be represented
in natural language (preferably English). The \emph{language} itself
could be represented via an ISO language code \cite{iso639}.
The \emph{source} could either focus on human-readability and, thus,
be a string describing the source, for example, using a citation in a
typical citation style. Otherwise, it could also be represented with
explicit metadata fields that contain the full bibliographic
information using, for example, the BibTeX data model
\cite{patashnik1988bibtexing}.
Relationships between contexts could be represented by directly
referring to other contexts using their (file) name(s).

The fields for each context could be represented as key-value pairs
and the metadata for all contexts can be represented as a list of
key-value pairs, where the keys are the names of contexts (i.e., their
file names) and the values are each a list of the key-value pairs of
the metadata for each context.
In principle, there exist many suitable data representations, for
example, RDF, XML, JSON, or YAML. Since we are aiming at a simple
human-editable solution, JSON and YAML seem to be good candidates.
Among those two, YAML has the additional benefit of using indentation
instead of brackets which might be easier to handle for humans not
trained in reading bracketed expressions.
Using YAML, the context from Example~\ref{ex:filename} could be
represented as follows:
\begin{minted}[breaklines]{yaml}
  livingbeings_en.cxt:
  - title: Living Beings and Water
  - source: "Ganter, B., & Wille, R. (1999). Formal Concept analysis. Springer, p. 18"
  - language: English
  - description: conditions different living beings need
\end{minted}

\paragraph{Where.} Like contexts, the metadata should also be
file-based. Initially, we consider one file describing all contexts to
be sufficient. The file shall reside in the highest level directory of
the repository, next to the \texttt{contexts} directory.
Another option would be to have one file with metadata for each
context. Apart from doubling the number of files this would make the
automatic retrieval of the information for several contexts more
difficult, since the name of the contexts need to be known in advance.
Furthermore, the metadata file shall also function as an index for the
contexts.
We are aware that the Burmeister format supports one line for a
comment. However, it is unclear whether and how the current tools
handle this comment and whether this line can be of any length. We
therefore refrain from explicitly using this comment line.

\subsubsection{Curation Policy}\label{sec:curation-policy}

As we have seen, there are different options on how to implement the
repository and many, often subtle, decisions have to be made. To
ensure consistency and ease of use, we consider it necessary to settle
some decisions beforehand in the form of guidelines for implementers
and contributors. We also think that such a \emph{curation policy}
will simplify and encourage contributions and -- if well-considered --
avoid some pitfalls (e.g., biases or compatibility problems).
The policy shall cover all aspects presented in
Section~\ref{sec:implementation} and, in principle, we consider all
decisions proposed there subject to discussion, for example, our
suggestions for file names.
Important questions are whether the policy should be regarded as
`rules' or as `guidelines' and how it should be enforced.
In~\cref{sec:working-group} we propose an approach on how to
settle these questions and how to implement such a policy.


\subsubsection{Other Aspects}\label{sec:other-aspects}
We briefly address some aspects we have not discussed so far:
\begin{itemize}
\item The repository should support keeping a history of changes (or
  version information) to be able to trace changes in the data and to
  enable referring to a specific version (which is important for
  replicable research). Using the git version control system automatically
  solves this problem.
\item Another issue are legacy aspects, for example, (limited) support
  for UTF-8 encoding in older tools, or differences in the choice of
  newline character(s) among operating systems. We are open for debate
  how to deal with such issues. However, this is probably not a
  practical problem either, as it can be solved dynamically by the git
  system.
\item We see a need for an accompanying support structure for the
  repository, for example, scripts to check the consistency and
  completeness of data and metadata, or to convert them from other
  formats.
\item To enable tools and libraries to access the contexts remotely, a
  remote API needs to be provided. Using a repository based on git
  typically provides this automatically. Specifically,  GitHub
  provides access using HTTPS and a REST-like API
  \cite{fielding2000architectural}. However, one might consider using
  a different transport protocol or providing a dedicated API (like
  DraCor~\cite{fischer2019programmable} does).
\end{itemize}

\section{Organization and Next Steps}\label{sec:organisation}
Our activity to create the FCA repository started with a post on the
fca-list mailing list\footnote{\url{fca-list@cs.uni-kassel.de}} in
February 2024 \cite{jaeschke2024request}.
Subsequently, we set up a git repository as part of the fcatools
organization on
GitHub\footnote{\url{https://github.com/fcatools/contexts}} and
published its web page on the domain
\texttt{fcarepository.org}.\footnote{\url{https://fcarepository.org/}}
%
Afterwards we uploaded twelve formal contexts together with their
metadata.  We describe the first steps we have taken to integrate the
repository into the FCA ecosystem in Section~\ref{sec:integration},
but first we state challenges we have observed or do anticipate
in~\cref{sec:challenges}. We envision to establish a working group,
as set out in~\cref{sec:working-group}. Finally, we want to limit the scope of
the repository in~\cref{sec:limits}.

\subsection{Challenges}\label{sec:challenges}

In establishing and maintaining the envisioned repository we foresee
several challenges that must be tackled.
%
Among the most crucial ones are \emph{acceptance and usage}
by the FCA community. We hope to set a good foundation with our
approach of early involvement of the community (e.g., our mailing list
post \cite{jaeschke2024request} and this paper), our suggestions for
implementation (e.g., prefer simple and robust approaches), and the
establishment of a working group to steer the future of the repository
(cf. the next section).
Another, but related challenge is the \emph{sustainability} of the
repository. This also affects technical issues, for example, the
choice of hosting service. We tackle this  by providing a dedicated
domain for the repository which simplifies migration to other
hosting services.
We aim to avoid a potential \emph{bias} in the selection and
description of the formal contexts by establishing procedures which
involve the community in the development of a curation policy.
Clearly, this list of challenges is not final (for further examples in
the context of OpenML see their
considerations~\cite{vanschoren2014openml}) but we consider those to
be the most important ones.

\subsection{Working Group}\label{sec:working-group}

We think that tackling the aforementioned challenges requires a
multi-stakeholder effort. Therefore, we propose to establish a
\emph{working group} which sustainably steers the extension, curation,
and dissemination of the repository.
The working group should be assembled by the FCA community to have
their trust and representation.  For practical purposes, this could
happen at the 2024 CONCEPTS conference or at a later community
meeting.
Concrete next steps the group should pursue include
\begin{itemize}
\item the community-driven establishment of a curation policy
  (cf. Section~\ref{sec:curation-policy}),
\item networking with other initiatives and collaboration (or
  federation) with related approaches (cf. Section~\ref{sec:related}),
\item development of accompanying resources, for example, workflows to
  check the integrity of contexts and metadata, and
\item establishing research data management measures, for example, the
  publication of regular snapshots of the repository in persistent
  research data repositories like Zenodo.
\end{itemize}

\subsection{Integration into the FCA Ecosystem}\label{sec:integration}

Our goal is that every relevant FCA software tool provides access to the
repository.
Similar to, for example, machine learning libraries\footnote{For
  example, scikit-learn has some basic data sets included which can be
  accessed with just one line of Python code: \mintinline{python}{iris
    = datasets.load_iris()}. Similarly, Seaborn's
  \mintinline{python}{load_dataset()} method loads data sets from the
  git repository \url{https://github.com/mwaskom/seaborn-data}.} we
propose that FCA software tools allow their users to load contexts from the
repository with just one line of code.
We have exemplarily implemented this in a
fork\footnote{\url{https://github.com/rjoberon/concepts/tree/example_datasets}}
of the Python library
\texttt{concepts},\footnote{\url{https://github.com/xflr6/concepts}} in which
it is possible to load the ``Living Beings'' context mentioned in
Example~\ref{ex:filename} as follows:
\begin{minted}{python}
  import concepts

  context = concepts.load_dataset('livingbeings_en')
\end{minted}

%
%
Upon approval of a pull
request\footnote{\url{https://github.com/xflr6/concepts/pull/24}} this
functionality will become available to all users of this Python
library.
Another pull
request\footnote{\url{https://github.com/tomhanika/conexp-clj/pull/141}}
will integrate similar functionality into conexp-clj.
To continue this effort, the working group shall reach out to
developers of other (recent) FCA software tools and libraries and support them in
integrating access to the repository.

\subsection{Limiting the Scope}\label{sec:limits}
We are fully aware that proper research data management comprises
more than just setting up a git repository. We consider our proposal
to be an important and concrete step towards actually having a
repository of formal contexts for FCA. We think it is better to have
an imperfect (in some respects) repository now than a perfect
repository later (or never).
Therefore, on purpose, we limit the scope of the repository as follows:
\begin{enumerate}[a)]
\item It is centered around formal contexts and there is no intention
  for a comprehensive or even complete modeling of all FCA data
  structures.
\item Its focus should be on the needs of the FCA community but not on
  the integration with other data modeling approaches, for example,
  linked data.
\item It should comprise contexts that are well-known or especially
  important for the FCA community. By well-known we mean that a
  context has been used in at least one published work or within a
  tutorial or in other educational capacities. We refrain from
  including purely generated contexts. Furthermore, no arbitrary data
  that (also) can be interpreted as contexts should be included.
\end{enumerate}


\section{Discussion}\label{sec:discussion}
In this paper we have outlined the first steps for (the only) real existing FCA repository. We have looked at approaches for repositories in other domains and drawn comparisons with the tools and services in the FCA ecosystem. We then carried out an in-depth analysis of the necessary parts and their implementation of the proposed FCA repository, which led to a number of preliminary but not firm design decisions. We expect the current state of the FCA repository to be initially resilient due to its anchoring in the Github service and its representation using the non-centralised versions control system git. In addition, choosing a domain for the repository that is independent of Github is a good choice for possibly changing the underlying hosting service later.

Nevertheless, there are a number of challenges to be met and many
(technical solutions) to be developed. These can only be overcome with
and through the FCA community. That is why -- with this work -- we want to invite them to take part in this endeavor.  The most crucial next step is for the community to find a reliable group of members who will take care of the repository and its further development.
We hope that this will be successful at one of the next scholarly meetings.





\printbibliography

\end{document}